\documentclass{svmult}
\usepackage{makeidx}         % allows index generation
\usepackage{graphicx}        % standard LaTeX graphics tool
\usepackage{multicol}        % used for the two-column index
\usepackage[bottom]{footmisc}% places footnotes at page bottom
\makeindex             % used for the subject index
\usepackage{amsmath,theorem}
\usepackage{amsfonts}
\usepackage{amssymb}
\usepackage{times,multirow,hhline,subfigure}
\usepackage{epsfig}
\usepackage{graphicx}
\usepackage[dcucite]{harvard}
\DeclareMathOperator*{\argmin}{\arg\min}
\usepackage{theorem}

\newtheorem{Algorithm}{Algorithm}

\begin{document}

\title*{Componentwise Least Squares Support Vector Machines}
\titlerunning{Componentwise LS-SVMs}
\author{K. Pelckmans\inst{1} \and
  I. Goethals \inst{1} \and
  J. De Brabanter \inst{1,2} \and
  J.A.K. Suykens\inst{1} \and
  B. De Moor \inst{1}}
\institute{KULeuven - ESAT - SCD/SISTA, \\
  Kasteelpark Arenberg 10, \\
  3001 Leuven (Heverlee), Belgium, \\
  \texttt{\{Kristiaan.Pelckmans,Johan.Suykens\}@esat.kuleuven.ac.be} \and
  Hogeschool KaHo Sint-Lieven (Associatie KULeuven), Departement Industrieel Ingenieur
}
\maketitle
                               
\begin{abstract}
  This chapter describes componentwise Least Squares Support Vector Machines (LS-SVMs) 
  for the estimation of additive models consisting of a sum of nonlinear components. 
  The primal-dual derivations characterizing LS-SVMs for the estimation of the additive 
  model result in a single set of linear equations with size growing in the number of data-points. 
  The derivation is elaborated for the classification as well as the regression case.
  Furthermore, different techniques are proposed to discover structure in the data 
  by looking for sparse components in the model based on dedicated regularization 
  schemes on the one hand and fusion of the componentwise LS-SVMs training with a 
  validation criterion on the other hand.
\end{abstract}

{\em keywords}: LS-SVMs, additive models, regularization, structure detection

%%%%%%%%%%%%%%%%%%%%%%
\section{Introduction}
%%%%%%%%%%%%%%%%%%%%%%

Non-linear classification and function approximation is an important topic of interest
with continuously growing research areas. Estimation techniques based on regularization
and kernel methods play an important role. We mention in this context 
smoothing splines \cite{wahba90}, regularization networks \cite{poggio90}, 
Gaussian processes \cite{mackay92}, Support Vector Machines (SVMs) 
\cite{vapnik98,cristianini00,schoelkopf02} and many more, see e.g. \cite{hastie01}.
SVMs and related methods have been introduced within the context of statistical 
learning theory and structural risk minimization. In the methods one solves convex 
optimization problems, typically quadratic programs. Least Squares Support Vector 
Machines (LS-SVMs)\footnote{http://www.esat.kuleuven.ac.be/sista/lssvmlab} 
\cite{suykens99,suykens02} are reformulations to standard SVMs which lead to 
solving linear KKT systems for classification tasks as well as regression.
In \cite{suykens02} LS-SVMs have been proposed as a class of kernel machines 
with primal-dual formulations in relation to kernel Fisher Discriminant Analysis (FDA), 
Ridge Regression (RR), Partial Least Squares (PLS), Principal Component Analysis (PCA), 
Canonical Correlation Analysis (CCA), recurrent networks and control. 
The dual problems for the static regression without bias term 
are closely related to Gaussian processes \cite{mackay92}, 
regularization networks \cite{poggio90} and Kriging \cite{cressie93}, 
while LS-SVMs rather take an optimization approach with primal-dual formulations
which have been exploited towards large scale problems and in developing robust versions.

Direct estimation of high dimensional nonlinear functions using 
a non-parametric technique without imposing restrictions faces 
the problem of the curse of dimensionality. Several attempts 
were made to overcome this obstacle, including projection 
pursuit regression \cite{friedmann81} and kernel methods for 
dimensionality reduction (KDR) \cite{fukumizu04}. Additive 
models are very useful for approximating high dimensional nonlinear 
functions \cite{stone85,hastie90}. These methods and their extensions have 
become one of the widely used nonparametric techniques as they 
offer a compromise between the somewhat conflicting requirements 
of flexibility, dimensionality and interpretability. Traditionally, 
splines are a common modeling technique \cite{wahba90} for additive 
models as e.g. in MARS (see e.g. \cite{hastie01}) or in combination 
with ANOVA \cite{neter74}. Additive models were brought further to 
the attention of the machine learning community by e.g. 
\cite{vapnik98,gunn02}. Estimation of the nonlinear components 
of an additive model is usually performed by the iterative 
{\em backfitting algorithm} \cite{hastie90} or a two-stage 
{\em marginal integration based} estimator \cite{linton95}. Although 
consistency of both is shown under certain conditions, important 
practical problems (number of iteration steps in the former) and more 
theoretical problems (the pilot estimator needed for the latter 
procedure is a too generally posed problem) are still left.

In this chapter we show how the primal-dual derivations characterizing LS-SVMs 
can be employed to formulate a straightforward solution to the estimation problem 
of additive models using convex optimization techniques for classification as well  
as regression problems. Apart from this one-shot optimal training algorithm, the 
chapter approaches the problem of structure detection in additive models 
\cite{hastie01,gunn02} by considering an appropriate regularization scheme leading 
to sparse components. The additive regularization (AReg) framework \cite{pelckmans03} is adopted 
to emulate effectively these schemes based on 2-norms, 1-norms and specialized penalization terms 
\cite{antoniadis01}. Furthermore, a validation criterion is considered to select relevant components. 
Classically, exhaustive search methods (or stepwise procedures)
are used which can be written as a combinatorial optimization problem. 
This chapter proposes a convex relaxation to the component selection problem.

This chapter is organized as follows. Section \ref{sect.clssvm} presents 
componentwise LS-SVM regressors and classifiers for efficient estimation 
of additive models and relates the result with ANOVA kernels and 
classical estimation procedures. Section \ref{sect.areg} introduces 
the additive regularization in this context and shows how to emulate 
dedicated regularization schemes in order to obtain sparse components.
Section \ref{sect.fusion} considers the problem of component 
selection based on a validation criterion. Section \ref{sect.ex} 
presents a number of examples.

%%%%%%%%%%%%%%%%%%%%%%%%%%%%%%%%%%
\section{Componentwise LS-SVMs and Primal-Dual Formulations}
%%%%%%%%%%%%%%%%%%%%%%%%%%%%%%%%%%
\label{sect.clssvm}

\subsection{The Additive Model Class}
%%%%%%%%%%%%%%%%%%%%%%%%%%%%%%%%%%%%%

Giving a training set defined as 
$\mathcal{D}_N = \{x_k,y_k\}_{k=1}^N \subset \mathbb{R}^D \times \mathbb{R}$ 
of size $N$ drawn i.i.d. from an unknown distribution $F_{XY}$ according to 
$y_k = f(x_k) + e_k$  where $f:\mathbb{R}^D \rightarrow \mathbb{R}$ 
is an unknown real-valued smooth function, $E[y_k|X=x_k] = f(x_k)$ 
and $e_1,\dots,e_N$ are uncorrelated random errors with $E\left[e_k|X = x_k\right]=0$, 
$E\left[(e_k)^2| X=x_k\right] = \sigma_e^2 < \infty$.
The $n$ data points of the validation set are denoted as 
$\mathcal{D}_n^{(v)} = \{x^{(v)}_j,y^{(v)}_j\}_{j=1}^{n}$.
The following vector notations are used throughout the text:
$X = \left(x_1,\dots,x_N\right) \in \mathbb{R}^{D \times N}$, 
$Y = \left(y_1,\dots,y_N\right)^T \in \mathbb{R}^{N}$,
$X^{(v)} = \left(x^{(v)}_1,\dots,x^{(v)}_n\right) \in \mathbb{R}^{D \times n}$ and 
$Y^{(v)} = \left(y^{(v)}_1,\dots,y^{(v)}_n\right)^T \in \mathbb{R}^{n}$.
The estimator of a regression function is difficult if the dimension $D$ is large.
One way to quantify this is the optimal minimax rate of convergence $N^{-2l/(2l+D)}$ 
for the estimation of an $l$ times differentiable regression function which converges to zero 
slowly if $D$ is large compared to $l$  \cite{stone82}. A possibility to overcome the curse 
of dimensionality is to impose additional structure on the regression function.
Although not needed in the derivation of the optimal solution, the input variables 
are assumed to be uncorrelated (see also {\em concurvity} \cite{hastie90}) in the applications. 

Let superscript $x^d \in \mathbb{R}$ denote the $d$-th component of an input vector $x\in\mathbb{R}^D$ 
for all $d=1,\dots,D$. Let for instance each component correspond with a different dimension of the input observations.
Assume that the function $f$ can be approximated arbitrarily well by 
a model having the following structure
\begin{equation}
  f(x) = \sum_{d=1}^D f^d(x^d) + b,
  \label{eq.clssvm}
\end{equation}
where $f^d:\mathbb{R} \rightarrow \mathbb{R}$ for all $d=1,\dots,D$ 
are unknown real-valued smooth functions and $b$  is an intercept term. 
The following vector notation is used: 
$X^d = \left(x^d_1,\dots,x^d_N\right) \in \mathbb{R}^{1 \times N}$ and 
$X^{(v)d} = \left(x^{(v)d}_1,\dots,x^{(v)d}_n\right) \in \mathbb{R}^{1 \times n}$.
The optimal rate of convergence for estimators based on this 
model is $N^{-2l/(2l+1)}$ which is independent of $D$ \cite{stone85}.
Most state-of-the-art estimation techniques for additive models can 
be divided into two approaches \cite{hastie01}:
\begin{itemize}
\item {\em Iterative approaches} use an iteration where in each step part of the unknown components 
  are fixed while optimizing the remaining components. This is motivated as:
  \begin{equation}
    \hat{f}^{d_1}(x^{d_1}_k) = y_k - e_k - \sum_{d_2 \neq d_1} \hat{f}^{d_2}(x^{d_2}_k),
  \end{equation}
  for all $k=1,\dots,N$ and $d_1 = 1,\dots,D$.
  Once the $N-1$ components of the second term are known, it becomes easy to 
  estimate the lefthandside. For a large class of linear smoothers, 
  such so-called backfitting algorithms are 
  equivalent to a Gauss-Seidel algorithm for solving a big ($ND \times ND$) 
  set of linear equations \cite{hastie01}. The backfitting algorithm 
  \cite{hastie90} is theoretically and practically well motivated.

\item {\em Two-stages marginalization approaches} construct in the first stage a 
  general black-box pilot estimator (as e.g. a Nadaraya-Watson kernel estimator) and 
  finally estimate the additive components by marginalizing (integrating out) 
  for each component the variation of the remaining components.

\end{itemize}

\subsection{Componentwise Least Squares Support Vector Machine Regressors}
%%%%%%%%%%%%%%%%%%%%%%%%%%%%%%%%%%%%%%%%%%%%%%%%%%%%%%%%%%%%%%%%%%%%%%%%%%%%%%%%
\label{subs.clssvmr}

At first, a primal-dual formulation is derived for componentwise LS-SVM regressors. 
The global model takes the form as in (\ref{eq.clssvm}) for any $x_\ast\in\mathbb{R}^D$
\begin{equation}
  f(x_\ast; w_d,b) = \sum_{d=1}^D f^d(x_\ast^d; w_d) + b = \sum_{d=1}^D {w_d}^T\varphi_d(x_\ast^d) + b.
  \label{eq.clssvmr.model}
\end{equation}
The individual components of an additive model based on LS-SVMs are written as 
$f^d(x^d; w_d) = w_d^T \varphi_d(x^d)$ in the primal space where $\varphi_d: 
\mathbb{R} \rightarrow \mathbb{R}^{n_{\varphi_d}}$ denotes a potentially infinite 
($n_{\varphi_d}=\infty$) dimensional feature map.
The regularized least squares cost function is given as \cite{suykens02}
\begin{multline}
  \min_{{w_d},b,e_k} \mathcal{J}_\gamma (w_d,e) =
  \frac{1}{2} \sum_{d=1}^D {w_d}^T w_d + \frac{\gamma}{2}\sum_{k=1}^N e_k^2 \\
  \text{ \ s.t. \ }  \sum_{d=1}^D {w_d}^T \varphi_d(x^d_k) + b + e_k = y_k, \text{ \ \ } k=1,\dots,N.
  \label{eq.clssvmr.cost}
\end{multline}
Note that the regularization constant $\gamma$ appears 
here as in classical Tikhonov regularization \cite{tikhonov77}. 
The Lagrangian of the constraint optimization problem becomes 
\begin{equation}
  \mathcal{L}_{\gamma}(w_d,b,e_k;\alpha_k) 
  = \frac{1}{2} \sum_{d=1}^D {w_d}^T w_d  + \frac{\gamma}{2} \sum_{k=1}^N e_k^2 
    - \sum_{k=1}^N \alpha_k ( \sum_{d=1}^D {w_d}^T \varphi_d(x_k^d) + b + e_k - y_k).
  \label{eq.clssvmr.lag}
\end{equation}
By taking the conditions for optimality 
$\partial \mathcal{L}_\gamma/\partial \alpha_k = 0$,  
$\partial \mathcal{L}_\gamma/\partial b  = 0$,  
$\partial \mathcal{L}_\gamma/\partial e_k = 0$ and
$\partial \mathcal{L}_\gamma/\partial w_d = 0 $
and application of the kernel trick $K^d(x^d_k,x^d_j) = \varphi_d(x^d_k)^T \varphi_d(x^d_j)$ with 
a positive definite (Mercer) kernel $K^d: \mathbb{R} \times\mathbb{R} \rightarrow \mathbb{R}$, 
one gets the following conditions for optimality
\begin{equation}
  \left\{\begin{array}{rlrl}
      y_k &= \sum_{d=1}^D {w_d}^T \varphi_d(x_k^d) + b + e_k,  & k=1,\dots,N & (a) \\
      e_k \gamma &= \alpha_k                              & k=1,\dots,N & (b) \\
      w_d &= \sum_{k=1}^N \alpha_k \varphi_d(x_k^d)        & d=1,\dots,D   & (c) \\
      0   &= \sum_{k=1}^N \alpha_k. && (d)
    \end{array}\right. 
  \label{eq.clssvmr.cond}
\end{equation}
Note that condition (\ref{eq.clssvmr.cond}.b) states that the elements of the solution vector $\alpha$
should be proportional to the errors. The dual problem is summarized in matrix notation as 
\begin{equation}
  \left[
    \begin{tabular}{c|c}
      $0$ & $1_{N}^{T}$ \\ \hline
      $1_{N}$ & $\Omega +I_{N}/\gamma$\\ 
    \end{tabular}
  \right]
  \left[ 
    \begin{tabular}{c}
      $b$ \\ \hline
      $\alpha$ \\ 
    \end{tabular}
  \right]
  = 
  \left[ 
    \begin{tabular}{c}
      $0$ \\ \hline
      $Y$ \\ 
    \end{tabular}
  \right],
  \label{eq.clssvmr.train}
\end{equation}
where $\Omega \in \mathbb{R}^{N \times N}$ with 
$\Omega = \sum_{d=1}^D \Omega^d$ and $\Omega^d_{kl} = K^d(x_k^d,x_l^d)$
for all $k,l=1,\dots,N$,  which is expressed in the dual variables 
$\hat\alpha$ instead of $\hat{w}$. A new point $x_\ast \in \mathbb{R}^D$ can be evaluated as 
\begin{equation}
  \hat{y}_\ast = \hat{f}^d(x_\ast; \hat\alpha,\hat{b}) 
  = \sum_{k=1}^N \hat\alpha_k \sum_{d=1}^D K^d(x_k^d,x_\ast^d) + \hat{b},
  \label{eq.clssvmr.eval}
\end{equation}
where $\hat\alpha$ and $\hat{b}$ is the solution to (\ref{eq.clssvmr.train}).
Simulating a validation datapoint $x_j$ for all $j=1,\dots,n$ by the $d$-th individual component
\begin{equation}
  \hat{y}^d_j = \hat{f}^d(x_j^d; \hat\alpha) = \sum_{k=1}^N \hat\alpha_k K^d(x_k^d,x_j^d),
  \label{eq.clssvmr.evald}
\end{equation}
which can be summarized as follows:
$\hat{Y}   = \left(\hat{y}_1,\dots,\hat{y}_N\right)^T \in \mathbb{R}^{N}$, 
$\hat{Y}^d$ $= \left(\hat{y}^d_1,\dots,\hat{y}^d_N\right)^T $ $\in \mathbb{R}^{N}$,
$\hat{Y}^{(v)} = \left(\hat{y}^{(v)}_1,\dots,\hat{y}^{(v)}_n\right)^T \in \mathbb{R}^{n}$ and 
$\hat{Y}^{(v)}_d = \left(\hat{y}^{(v)d}_1,\dots,\hat{y}^{(v)d}_n\right)^T \in \mathbb{R}^{n}$. 

\begin{figure}
   \begin{center}
     \scalebox{.6}{\epsfig{figure=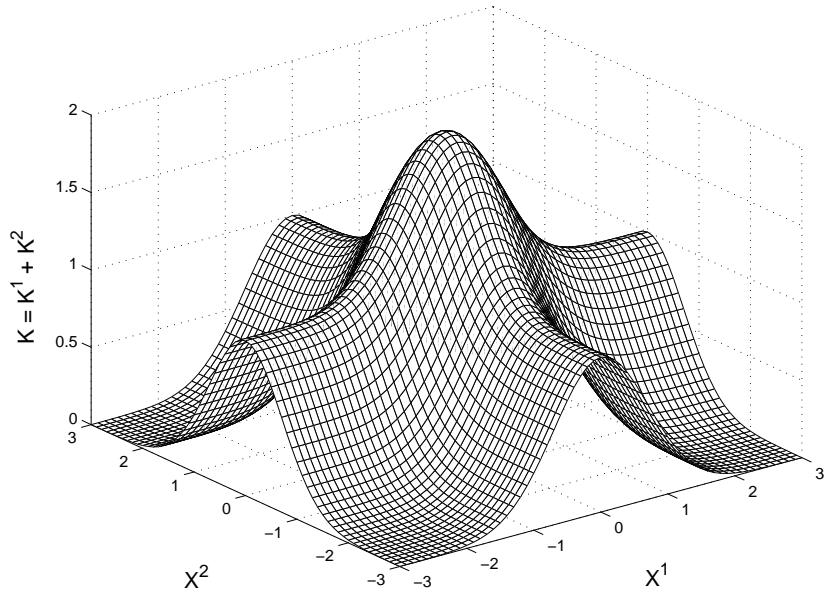}}
   \end{center}
   \caption{\sl The two dimensional componentwise Radial Basis Function (RBF) 
     kernel for componentwise LS-SVMs takes the form 
     $K(x_k,x_l) =K^1(x^1_k,x^1_l) + K^2(x^2_k,x^2_l)$ as displayed.
     The standard RBF kernel takes the form $K(x_k,x_l) = \exp(-\|x_k - x_l\|^2_2/\sigma^2)$ 
     with $\sigma \in\mathbb{R}^+_0$ an appropriately chosen bandwidth.}
   \label{fig.arbf}
 \end{figure}

\noindent{\bf Remarks:}
\begin{itemize}
\item Note that the componentwise LS-SVM regressor can be written as a linear 
  smoothing matrix \cite{suykens02}:
  \begin{equation}
    \hat{Y} =  S_\gamma Y. 
    \label{eq.clssvmr.linear}
  \end{equation}
  For notational convenience, the bias term is omitted from this description.
  The smoother matrix $S_\gamma \in \mathbb{R}^{N \times N}$ becomes
  \begin{equation}
    S_\gamma = \Omega \left(\Omega + I_N\frac{1}{\gamma}\right)^{-1}. 
    \label{eq.clssvmr.smoother}
  \end{equation}

\item The set of linear equations (\ref{eq.clssvmr.train}) 
  corresponds with a classical LS-SVM regressor 
  where a modified kernel is used 
  \begin{equation}
    K(x_k,x_j) = \sum_{d=1}^D K^d(x^d_k,x_j^d).
    \label{eq.clssvmr.ak}
  \end{equation}
  Figure \ref{fig.arbf} shows the modified kernel in case a one dimensional 
  Radial Basis Function (RBF) kernel is used for all $D$ (in the example, $D=2$) components. 
  This observation implies that componentwise LS-SVMs inherit results obtained 
  for classical LS-SVMs and kernel methods in general. From a practical point of view, 
  the previous kernels (and a fortiori componentwise kernel models) result in the 
  same algorithms as considered in the ANOVA kernel decompositions as in \cite{vapnik98,gunn02}. 
  \begin{equation}
    K(x_k,x_j) 
    = \sum_{d=1}^D K^d(x^d_k,x_j^d) 
    + \sum_{d_1 \neq d_2} K^{d_1 d_2}\left((x_k^{d_1}, x^{d_2}_k)^T,(x_j^{d_1}, x_j^{d_2})^T\right) 
    + \dots,
    \label{eq.clssvmr.anova}
  \end{equation}
  where the componentwise LS-SVMs only consider the first term in this expansion.
  The described derivation as such bridges the gap between the estimation of additive models 
  and the use of ANOVA kernels.

\end{itemize}

\subsection{Componentwise Least Squares Support Vector Machine Classifiers}
%%%%%%%%%%%%%%%%%%%%%%%%%%%%%%%%%%%%%%%%%%%%%%%%%%%%%%%%%%%%%%%%%%%%%%%%%%%
\label{subs.clssvmc}

In the case of classification, let $y_k,y_j^{(v)} \in \{-1,1\}$ for all $k=1,\dots,N$ and $j=1,\dots,n$.
The analogous derivation of the componentwise LS-SVM classifier is briefly reviewed.
The following model is considered for modeling the data
\begin{equation}
  f(x) = {\rm sign}\left(\sum_{d=1}^D f^d(x^d) + b \right),
  \label{eq.clssvmc}
\end{equation}
where again the individual components of the additive model based on LS-SVMs are given as 
$f^d(x^d) = {w_d}^T \varphi_d(x^d)$ in the primal space where 
$\varphi_d: \mathbb{R} \rightarrow \mathbb{R}^{n_{\varphi_d}}$ denotes a potentially 
infinite ($n_{\varphi_d}=\infty$) dimensional feature map. The regularized least squares 
cost function is given as \cite{suykens99,suykens02}
\begin{multline}
  \min_{w_d,b,e_k} \mathcal{J}_\gamma (w_d,e) =
  \frac{1}{2} \sum_{d=1}^D {w_d}^T w_d + \frac{\gamma}{2}\sum_{k=1}^N e_k^2 \\
  \text{ \ s.t. \ }  y_k \left( \sum_{d=1}^D {w_d}^T \varphi_d(x^d_k) + b \right) = 1- e_k, 
  \text{ \ \ } k=1,\dots,N,
  \label{eq.clssvmc.cost}
\end{multline}
where $e_k$ are so-called slack-variables for all $k=1,\dots,N$.
After construction of the Lagrangian and taking the conditions for optimality, one obtains 
the following set of linear equations (see e.g. \cite{suykens02}):
\begin{equation}
  \left[
    \begin{tabular}{c|c}
      $0$ & $Y^T$ \\ \hline
      $Y$ & $\Omega_y +I_{N}/\gamma$\\ 
    \end{tabular}
  \right]
  \left[ 
    \begin{tabular}{c}
      $b$ \\ \hline
      $\alpha$ \\ 
    \end{tabular}
  \right]
  = 
  \left[ 
    \begin{tabular}{c}
      $0$ \\ \hline
      $1_N$ \\ 
    \end{tabular}
  \right],
  \label{eq.clssvmc.train}
\end{equation}
where $\Omega_y \in \mathbb{R}^{N \times N}$ with 
$\Omega_y = \sum_{d=1}^D \Omega^d_y \in\mathbb{R}^{N \times N}$
and $\Omega^d_{y,kl} = y_k y_l K^d(x_k^d,x_l^d)$.
New data points $x_\ast \in \mathbb{R}^D$ can be evaluated as 
\begin{equation}
  \hat{y}_\ast = {\rm sign}\left( \sum_{k=1}^N \hat\alpha_k y_k \sum_{d=1}^D K^d(x_k^d,x_\ast^d) + \hat{b} \right).
  \label{eq.clssvmc.eval}
\end{equation}
In the remainder of this text, only the regression case is considered. 
The classification case can be derived straightforwardly along the lines.

%%%%%%%%%%%%%%%%%%%%%%%%%%%%%%%%%%%%%%%%%%%%%%%%%%%%%%%%%%%%%%%%%%%%%%%%
\section{Regularizing for Sparse Components via Additive Regularization}
%%%%%%%%%%%%%%%%%%%%%%%%%%%%%%%%%%%%%%%%%%%%%%%%%%%%%%%%%%%%%%%%%%%%%%%%
\label{sect.areg}

A regularization method fixes a priori the answer to the ill-conditioned 
(or ill-defined) nature of the inverse problem. The classical Tikhonov 
regularization scheme \cite{tikhonov77} states the answer in terms of the norm 
of the solution. The formulation of the additive regularization (AReg) 
framework \cite{pelckmans03} made it possible to impose alternative 
answers to the ill-conditioning of the problem at hand.
We refer to this AReg level as {\em substrate LS-SVMs}.
An appropriate regularization scheme for additive models 
is to favor solutions using the smallest number of components 
to explain the data as much as possible. 
In this paper, we use the somewhat relaxed condition of {\em sparse} 
components to select appropriate components instead of the 
more general problem of input (or component) selection.

\subsection{Level 1: Componentwise LS-SVM Substrate}
%%%%%%%%%%%%%%%%%%%%%%%%%%%%%%%%%%%%%%%%%%%%%%%%%%%%
\label{subs.areg}

\begin{figure}
  \begin{center}
    \scalebox{.4}{\epsfig{figure=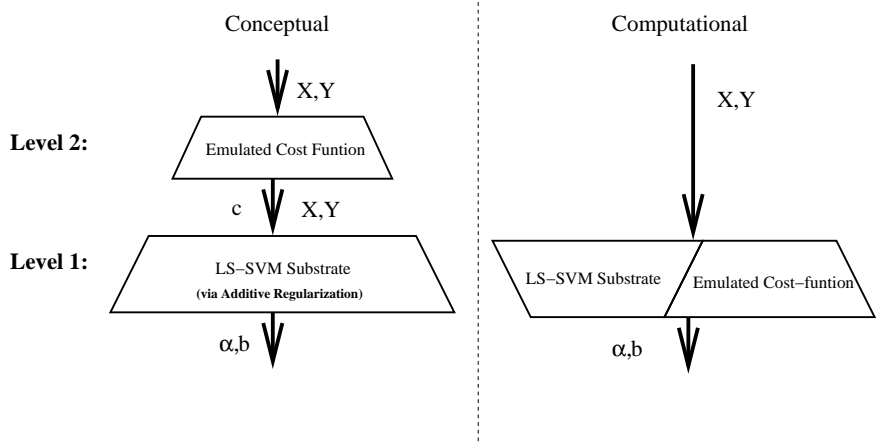}}
  \end{center}
  \caption{\sl Graphical representation of the additive regularization framework used 
    for emulating other loss functions and regularization schemes. 
    Conceptually, one differentiates between the newly specified cost 
    function and the LS-SVM substrate, while computationally both are computed simultanously.}
  \label{fig.substrate}
\end{figure}
Using the Additive Regularization (AReg) scheme for componentwise LS-SVM regressors results into
the following modified cost function:
\begin{multline}
  \min_{w_d,b,e_k} \mathcal{J}_c (w_d,e) =
  \frac{1}{2} \sum_{d=1}^D {w_d}^T w_d + \frac{1}{2}\sum_{k=1}^N (e_k - c_k)^2 \\
  \text{ \ s.t. \ }  \sum_{d=1}^D {w_d}^T \varphi_d(x^d_k) + b + e_k = y_k, \text{ \ \ } k=1,\dots,N,
  \label{eq.areg.cost}
\end{multline}
where $c_k \in \mathbb{R}$ for all $k=1,\dots,N$. 
Let $c = (c_1,\dots,c_N)^T \in \mathbb{R}^N$.
After constructing the Lagrangian and taking the conditions for optimality, one obtains 
the following set of linear equations, see \cite{pelckmans03}:
\begin{equation}
  \left[
    \begin{tabular}{c|c}
      $0$ & $1_{N}^{T}$ \\ \hline
      $1_{N}$ & $\Omega +I_{N}$\\ 
    \end{tabular}
  \right]
  \left[ 
    \begin{tabular}{c}
      $b$ \\ \hline
      $\alpha$ \\ 
    \end{tabular}
  \right]
  +
  \left[ 
    \begin{tabular}{c}
      $0$ \\ \hline
      $c$ \\ 
    \end{tabular}
  \right]
  = 
  \left[ 
    \begin{tabular}{c}
      $0$ \\ \hline
      $Y$ \\ 
    \end{tabular}
  \right]
  \label{eq.areg.train}
\end{equation}
and $e = \alpha +c \in \mathbb{R}^N$.
Given a regularization constant vector $c$, the unique solution follows
immediately from this set of linear equations. 

However, as this scheme is too general for practical implementation, $c$ should be limited in an appropriate way
by imposing for example constraints corresponding with certain model assumptions or a specified cost function.
Consider for a moment the conditions for optimality of the componentwise LS-SVM regressor
using a regularization term as in ridge regression, one can see that equation (\ref{eq.clssvmr.train}) 
corresponds with (\ref{eq.areg.train}) if $\gamma^{-1}\alpha = \alpha+c$ for given $\gamma$.
Once an appropriate $c$ is found which satisfies the constraints, 
it can be plugged in into the LS-SVM substrate (\ref{eq.areg.train}). 
It turns out that one can omit this conceptual second stage  in the computations 
by elimination of the variable $c$ in the constrained optimization problem (see Figure \ref{fig.substrate}).

\begin{figure}
   \begin{center}
     \scalebox{.4}{\epsfig{figure=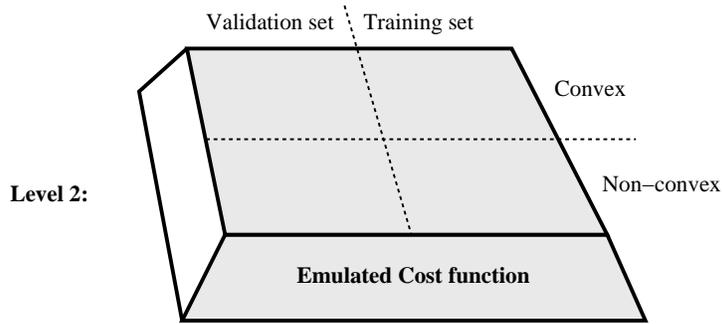}}
   \end{center}
   \caption{\sl The level 2 cost functions of Figure \ref{fig.substrate} on the conceptual level 
     can take different forms based on validation performance or trainings error. While some 
     will result in convex tuning procedures, other may loose this property depending on the chosen
     cost function on the second level. }
   \label{fig.emulated}
 \end{figure}

Alternatively, a measure corresponding with a (penalized) cost function can be used which fulfills
the role of model selection in a broad sense. A variety of such explicit or implicit 
limitations can be {\em emulated} based on different criteria (see Figure \ref{fig.emulated}).

\subsection{Level 2: Emulating an $L_1$ based Component Regularization Scheme {\em (Convex)}}
%%%%%%%%%%%%%%%%%%%%%%%%%%%%%%%%%%%%%%%%%%%%%%%%%%%%%%%%%%%%%%%
\label{subs.1norm}

We now study how to obtain sparse components by 
considering a dedicated regularization scheme. 
The LS-SVM substrate technique is used to emulate the proposed scheme as 
primal-dual derivations (see e.g.  Subsection \ref{subs.clssvmr}) 
are not straightforward anymore.

Let $\hat{Y}^d \in \mathbb{R}^N$ denote the estimated training outputs 
of the $d$-th submodel $f^d$ as in (\ref{eq.clssvmr.evald}).
The component based regularization scheme can be translated 
as the following constrained optimization problem where the conditions 
for optimality (\ref{eq.areg.cost}) as summarized in (\ref{eq.areg.train}) 
are to be satisfied exactly (after elimination of $w$)
\begin{multline}
  \min_{c,\hat{Y}^d,e_k; \alpha,b} \mathcal{J}_\xi (\hat{Y}^d,e_k) =
  \frac{1}{2} \sum_{d=1}^D \|\hat{Y}^d\|_1 + \frac{\xi}{2}\sum_{k=1}^N e_k^2 \\
  \text{ \ s.t. \ }  
  \left\{\begin{array}{l}
      1_N^T \alpha = 0, \\
      \Omega \,\alpha + 1_N^T b + \alpha+c = Y, \\
      \Omega^d \,\alpha = \hat{Y}^d, \ \ \ \forall d = 1,\dots,D \\
      \alpha +c = e, 
  \end{array}\right. 
  \label{eq.aregam.cost}
\end{multline}
where the use of the robust $L_1$ norm can be justified as in general no assumptions are imposed 
on the distribution of the elements of $\hat{Y}^d$.
By elimination of $c$ using the equality $e = \alpha + c$, this problem can be written as follows
\begin{multline}
  \min_{\hat{Y}^d,e_k; \alpha,b} \mathcal{J}_\xi (\hat{Y}^d,e_k) =
  \frac{1}{2} \sum_{d=1}^D \|\hat{Y}^d\|_1 + \frac{\xi}{2}\sum_{k=1}^N e_k^2 \\
  \text{ \ s.t. \ }  
  \left[
    \begin{tabular}{c|c}
      $0$ & $1_{N}^{T}$ \\ \hline
      $1_{N}$ & $\Omega$\\ \hline 
      $0_{N}$ & $\Omega^1$\\ 
      $\vdots$ & $\vdots$\\ 
      $0_{N}$ & $\Omega^d$\\ 
    \end{tabular}
  \right]
  \left[ 
    \begin{tabular}{c}
      $b$ \\ \hline
      $\alpha$ \\ 
    \end{tabular}
  \right]
  +
  \left[ 
    \begin{tabular}{c}
      $0$ \\ \hline
      $e$ \\ \hline
      $\hat{Y}^1$ \\ 
      $\vdots$ \\ 
      $\hat{Y}^D$ \\ 
    \end{tabular}
  \right]
  = 
  \left[ 
    \begin{tabular}{c}
      $0$ \\ \hline
      $Y$ \\ \hline
      $0_N$ \\ 
      $\vdots$ \\ 
      $0_N$ \\ 
    \end{tabular}
  \right].
  \label{eq.aregam.costb}
\end{multline}
This convex constrained optimization problem can be solved as a quadratic programming problem.
%A more efficient algorithmic approach is described in Subsection \ref{subs.wgna}.
As a consequence of the use of the $L_1$ norm, often sparse components ($\|\hat{Y}^d\|_1 = 0$)
are obtained, in a similar way as sparse variables of LASSO or sparse datapoints in SVM \cite{hastie01,vapnik98}. 
An important difference is that the estimated outputs are used for regularization purposes 
instead of the solution vector. It is good practice to omit sparse components on the 
training dataset from  simulation:
\begin{equation}
  \hat{f}(x_\ast; \hat\alpha,\hat{b}) 
  = \sum_{i=1}^N \hat\alpha_i \sum_{d \in\mathcal{S}_D} K^d (x_i^d,x_\ast^d) + \hat{b},
  \label{eq.aregam.sim}
\end{equation}
where $\mathcal{S}_D = \{d | \hat\alpha^T \Omega^d \hat\alpha \neq 0\}$.

Using the $L_2$ norm $\sum_{d=1}^D \|\hat{Y}^d\|_2^2$ instead leads to a much simpler 
optimization problem, but additional assumptions (Gaussianity) are needed on the distribution
of the elements of $\hat{Y}^d$. Moreover, the component selection has to resort on 
a significance test instead of the sparsity resulting from (\ref{eq.aregam.costb}). 
A practical algorithm is proposed in Subsection \ref{subs.wgna} that uses 
an iteration of $L_2$ norm based optimizations in order to calculate the 
optimum of the proposed regularized cost function.

\subsection{Level 2 bis: Emulating a Smoothly Thresholding Penalty Function {\em (Non-convex)}}
%%%%%%%%%%%%%%%%%%%%%%%%%%%%%%%%%%%%%%%%%%%%%%%%%%%%%%%%%%%%%%%%
\label{subs.oracle}

\begin{figure}
  \centering
  \begin{tabular}{ccc}\\
    \subfigure[]{\scalebox{.23}{\epsfig{figure=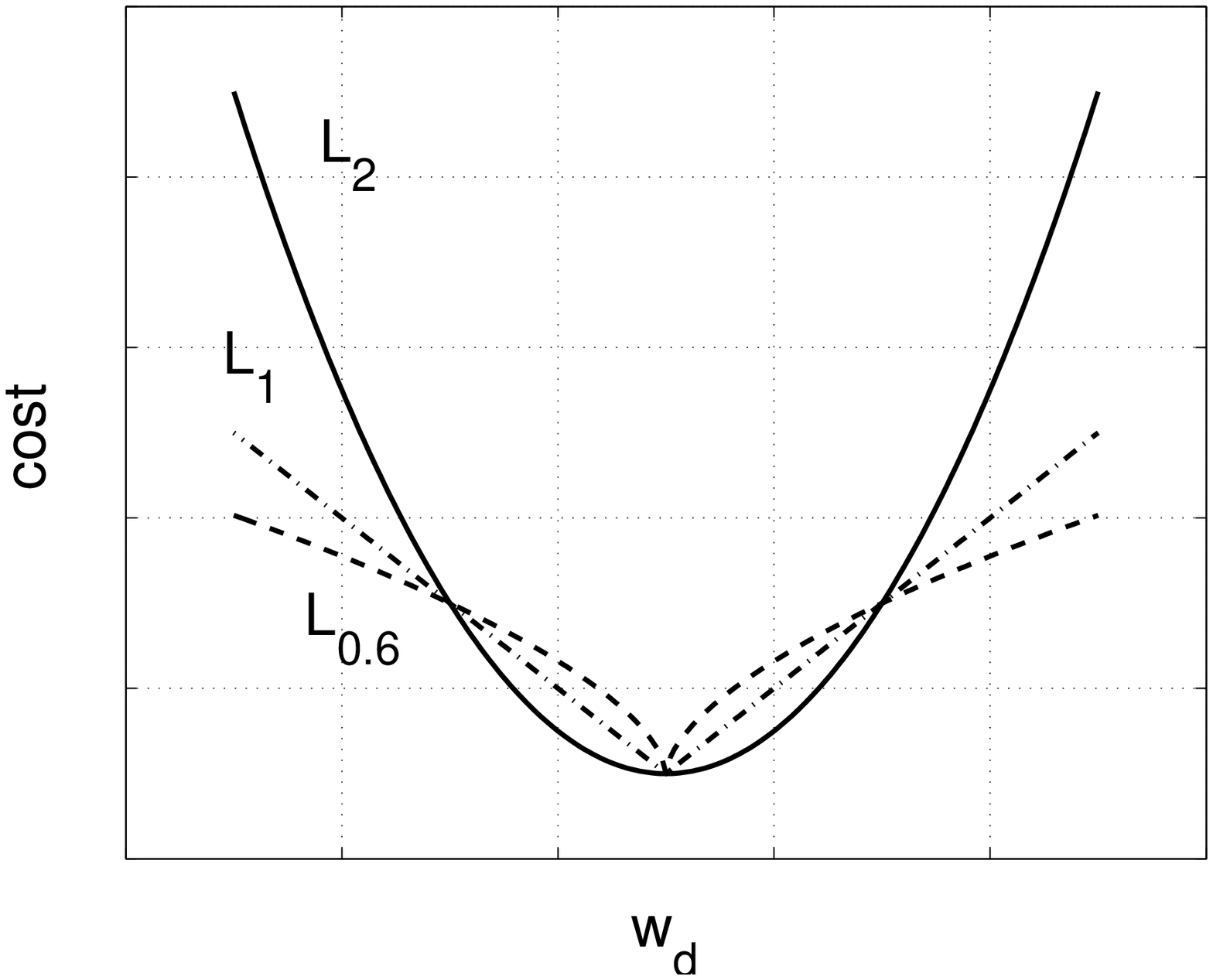}}} &
    \subfigure[]{\scalebox{.23}{\epsfig{figure=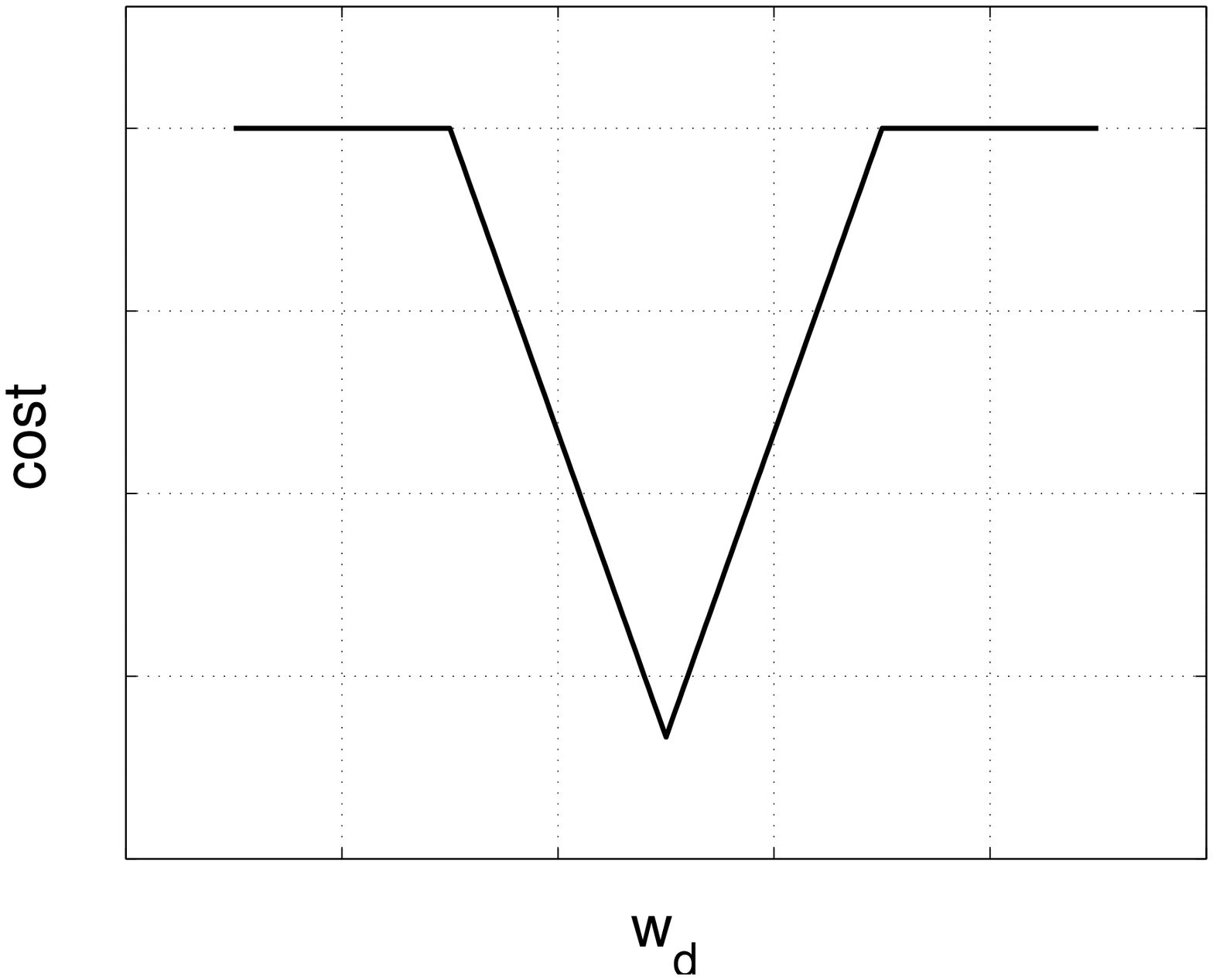}}}& 
    \subfigure[]{\scalebox{.23}{\epsfig{figure=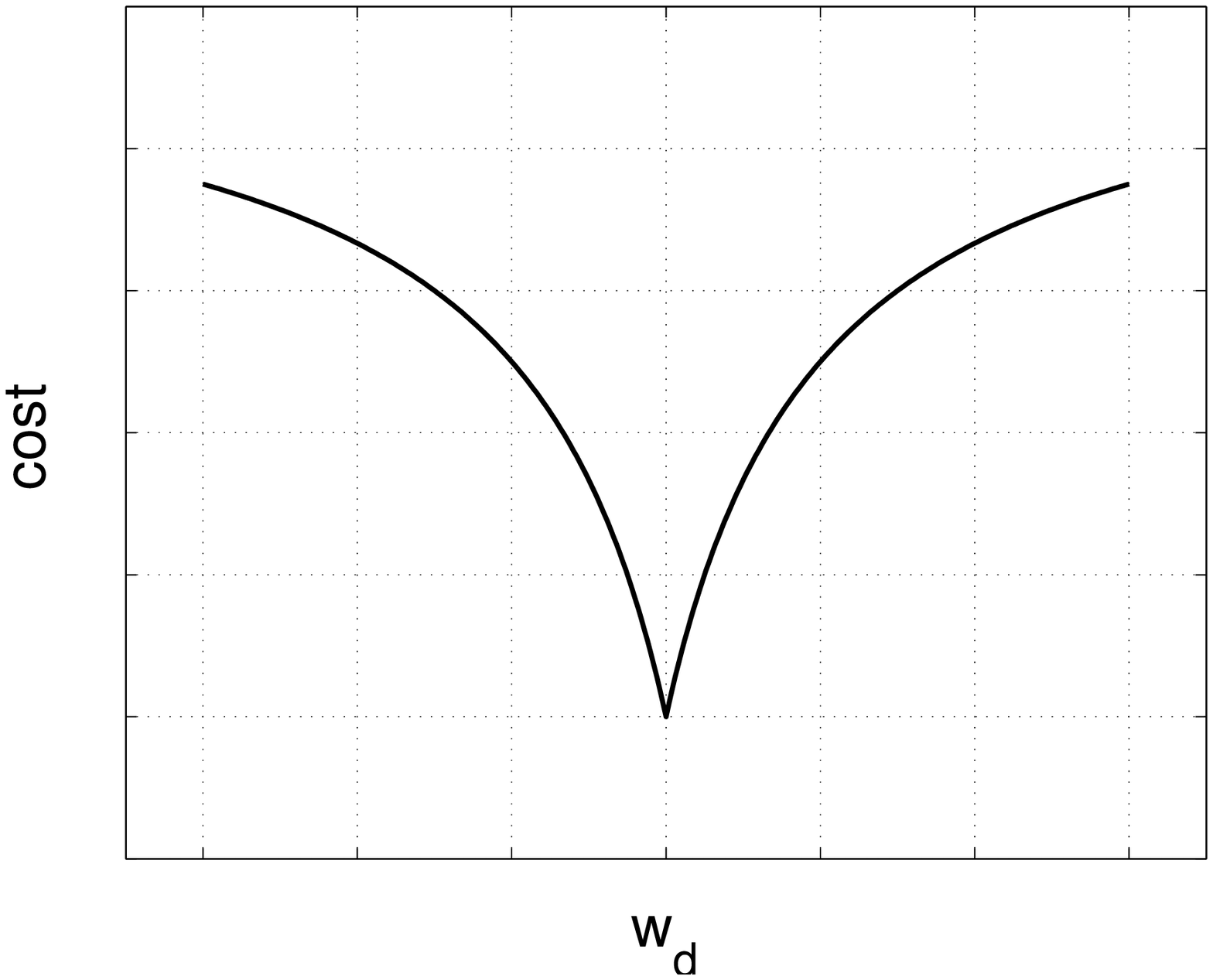}}} 
    \end{tabular}
  \caption{\sl Typical penalty functions:
    {\bf (a)} the $L_p$ penalty family for $p=2,1$ and $0.6$, 
    {\bf (b)} hard thresholding penalty function and 
    {\bf (c)} the transformed $L_1$ penalty function.}
  \label{fig.penalty}
\end{figure}

This subsection considers extensions to classical formulations towards the use 
of dedicated regularization schemes for sparsifying components.
Consider the componentwise regularized least squares cost function defined as 
\begin{equation}
  \mathcal{J}_\lambda (w_d,e) 
  = \frac{\lambda}{2} \sum_{d=1}^D \ell(w_d) + \frac{1}{2} \sum_{k=1}^N e^2_k,
  \label{eq.rls}
\end{equation}
where $\ell(w_d)$ is a penalty function and $\lambda \in\mathbb{R}_0+$ 
acts as  a regularization parameter.  We denote $\lambda \ell(\cdot)$ by $\ell_\lambda(\cdot)$, 
so it may depend on $\lambda$. Examples of penalty functions include:
\begin{itemize}
\item The $L_p$ penalty function $\ell_\lambda^p(w_d) = \lambda \|w_d\|^p_p$ leads to a bridge regression 
  \cite{frank93,fu98}. It is known that the $L_2$ penalty function $p=2$ results in 
  the ridge regression. For the $L_1$ penalty function the solution is the soft thresholding rule 
  \cite{donoho94}. LASSO, as proposed by \cite{tibshirani96,tibshirani97}, is the penalized 
  least squares estimate using the $L_1$ penalty function (see Figure \ref{fig.penalty}.a).
  
\item Let the indicator function $I_{\{x \in\mathcal{A}\}} = 1$ if $x \in\mathcal{A}$ 
  for a specified set $\mathcal{A}$ and 0 otherwise. When the penalty function is given by 
  $\ell_\lambda(w_d) = \lambda^2 - (\|w_d\|_1-\lambda)^2 I_{\{\|w_d\|_1<\lambda\}}$ 
  (see Figure \ref{fig.penalty}.b), the solution is a hard-thresholding rule \cite{antoniadis97}. 

\end{itemize}
The $L_p$ and the hard thresholding penalty functions do not simultaneously
satisfy the mathematical conditions for unbiasedness, sparsity and continuity \cite{fan01}.
The hard thresholding has a discontinuous cost surface. 
The only continuous cost surface (defined as the cost function associated with the solution space) 
with a thresholding rule in the $L_p$-family is the $L_1$ penalty function, 
but the resulting estimator is shifted by a constant $\lambda$. 
To avoid these drawbacks, \cite{nikolova99} suggests the 
penalty function defined as 
\begin{equation}
  \ell_\lambda^a(w_d) = \frac{\lambda a\|w_d\|_1}{1+a\|w_d\|_1},
  \label{eq.scad}
\end{equation}
with $a \in\mathbb{R}_0^+$.
This penalty function behaves quite similarly as the Smoothly Clipped Absolute Deviation (SCAD) 
penalty function as suggested by \cite{fan97}. The Smoothly Thresholding Penalty (TTP) function (\ref{eq.scad})
improves the properties of the $L_1$ penalty function and the hard thresholding penalty function 
(see Figure \ref{fig.penalty}.c), see \cite{antoniadis01}.
The unknowns $a$ and $\lambda$ act as regularization parameters. A plausible value for $a$ 
was derived in \cite{nikolova99,antoniadis01} as $a= 3.7$.
The transformed $L_1$ penalty function satisfies the oracle inequalities \cite{donoho94}.
One can plugin the described semi-norm $\ell_\lambda^a(\cdot)$ to improve the 
component based regularization scheme (\ref{eq.aregam.cost}). 
Again, the additive regularization scheme is used for the emulation of this scheme
\begin{multline}
  \min_{c,\hat{Y}^d,e_k; \alpha,b} \mathcal{J}_\lambda (\hat{Y}^d,e_k) =
  \frac{1}{2} \sum_{d=1}^D \ell_\lambda^a(\hat{Y}^d) + \frac{1}{2}\sum_{k=1}^N e_k^2 \\
  \text{ \ s.t. \ }  
  \left\{\begin{array}{l}
      1_N^T \alpha = 0, \\
      \Omega \,\alpha + 1_N^T b + \alpha+c = Y, \\
      \Omega^d \,\alpha = \hat{Y}^d, \ \ \ \forall d = 1,\dots,D \\
      \alpha +c = e,
  \end{array}\right. 
  \label{eq.aregam.cost2}
\end{multline}
which becomes non-convex but can be solved using an iterative scheme as 
explained later in Subsection \ref{subs.wgna}.

\section{Fusion of Componentwise LS-SVMs and Validation}
%%%%%%%%%%%%%%%%%%%%%%%%%%%%%%%%%%%%%%%%%%%%%%%%%%%%%%%%
\label{sect.fusion}

This section investigates how one can tune the componentwise LS-SVMs 
with respect to a validation criterion in order to improve the generalization performance 
of the final model. As proposed in \cite{pelckmans03}, fusion of training and validation 
levels can be investigated from an optimization point of view, while conceptually they are
to be considered at different levels.

\subsection{Fusion of Componentwise LS-SVMs and Validation for Regularization Constant Tuning}

For this purpose, the fusion argument as introduced in \cite{pelckmans03} 
is briefly revised in relation to regularization parameter tuning.
The estimator of the LS-SVM regressor on the training data for 
a fixed value $\gamma$ is given as (\ref{eq.clssvmr.cost})
\begin{equation}
  {\bf Level \ 1: \ \ } 
  (\hat{w},\hat{b}) = \argmin_{w,b,e} \mathcal{J}_\gamma(w,e)
  \mbox{ \ \ s.t. (\ref{eq.clssvmr.cost}) holds,}
  \label{eq.fusion.lssvm}
\end{equation}
which results into solving a linear set of equations (\ref{eq.clssvmr.train}) after substitution of $w$
by Lagrange multipliers $\alpha$. Tuning the regularization parameter by using a 
validation criterion gives the following estimator
\begin{equation}
  {\bf Level \ 2: \ \ }
  \hat\gamma  = \argmin_\gamma \sum_{j=1}^n \left( f(x_j ;\hat\alpha,\hat{b}) - y_j\right)^2
  \mbox{ \ \ with \ \ } (\hat\alpha,\hat{b}) = \argmin_{\alpha,b} \mathcal{J}_\gamma
 \label{eq.fusion.tuning}
\end{equation}
satisfying again (\ref{eq.clssvmr.cost}). Using the conditions for optimality 
(\ref{eq.clssvmr.train}) and eliminating $w$ and $e$ 
\begin{equation}
  {\bf Fusion: \ \ }
  (\hat\gamma,\hat\alpha,\hat{b}) = \argmin_{\gamma,\alpha,b} \sum_{j=1}^n \left( f(x_j ;\alpha,b) - y_j\right)^2
  \mbox{ \ \ s.t. \ \ (\ref{eq.clssvmr.train}) holds,}\hspace{+10mm}
  \label{eq.fusion.fusion}
\end{equation}
which is referred to as fusion. The resulting optimization problem 
was noted to be non-convex as the set of optimal solutions $w$ (or dual $\alpha$'s) 
corresponding with a $\gamma >0$ is non-convex. To overcome 
this problem, a re-parameterization of the trade-off was proposed leading 
to the additive regularization scheme. At the cost of overparameterizing 
the trade-off, convexity is obtained. To circumvent this drawback, 
different ways to restrict explicitly or implicitly the (effective) degrees 
of freedom of the regularization scheme $c \in\mathcal{A} \subset \mathbb{R}^N$ 
were proposed while retaining convexity (\cite{pelckmans03}). The convex 
problem resulting from additive regularization is
\begin{equation}
  {\bf Fusion: \ \ }
  (\hat{c},\hat\alpha,\hat{b}) 
  = \argmin_{c\in\mathcal{A},\alpha,b} \sum_{j=1}^n \left( f(x_j ;\alpha,b) - y_j\right)^2
  \mbox{ \ \ s.t. \ \ (\ref{eq.areg.train}) holds}, \hspace{+10mm}
  \label{eq.areg.fusion2}
\end{equation}
and can be solved efficiently as a convex constrained optimization problem if $\mathcal{A}$ is a convex set, 
resulting immediately in the optimal regularization trade-off and model parameters \cite{boyd04}.

\subsection{Fusion for Component Selection using the Additive Regularization Scheme}
%%%%%%%%%%%%%%%%%%%%%%%%%%%%%%%%%%%%%%%%%%%%%%%%%%%%%%%%%%%%%%%%%%%%%%%%%%%%%%%%%%%%

One possible relaxed version of the component selection problem goes as follows:
{\em Investigate whether it is plausible to drive the components on the validation 
set to zero without too large modifications on the global training solution}.
This is translated as the following cost function much in the spirit of (\ref{eq.aregam.cost}).
Let $\Omega^{(v)}$ denote $\sum_{d=1}^D \Omega^{(v)d} \in\mathbb{R}^{n\times N}$ 
and $\Omega^{(v)d}_{jk} = K^d(x^{(v)d}_j,x_k^d)$ for all $j=1,\dots,n$ and $k=1,\dots,N$.
\begin{multline}
  (\hat{c},\hat{Y}^{(v)d},\hat{w}_d,\hat{e},\hat\alpha,\hat{b}) 
  = \argmin_{c,\hat{Y}^d,\hat{Y}^{(v)d},e,\alpha,b} 
              \frac{1}{2}     \sum_{d=1}^D \|\hat{Y}^{(v)d}\|_1 
            + \frac{1}{2}     \sum_{d=1}^D \|\hat{Y}^{d}\|_1 
            + \frac{\xi}{2}\sum_{k=1}^N e_k^2 \\
  \text{ \ s.t. \ }  
  \left\{\begin{array}{l}
      1_N^T \alpha = 0\\
      \alpha +c = e \\
      \Omega \,\alpha + 1_N b + \alpha + c = Y \\
      \Omega^{d} \,\alpha = \hat{Y}^{d}, \ \ \ \forall d = 1,\dots,D, \\
      \Omega^{(v)d} \,\alpha = \hat{Y}^{(v)d}, \ \ \ \forall d = 1,\dots,D,
  \end{array}\right. 
  \label{eq.fusion1.cost}
\end{multline}
where the equality constraints consist of the conditions for optimality 
of (\ref{eq.areg.train}) and the evaluation of the validation set on the individual components.
Again, this convex problem can be solved as a quadratic programming problem.

\subsection{Fusion for Component Selection using Componentwise Regularized LS-SVMs}
%%%%%%%%%%%%%%%%%%%%%%%%%%%%%%%%%%%%%%%%%%%%%%%%%%%%%%%%%%%%%%%%%%%%%%%%%%%%%%%%%%%%

We proceed by considering the following primal cost function for a 
fixed but strictly positive $\eta = (\eta_1,\dots,\eta_D)^T \in (\mathbb{R}_0^+)^D$
\begin{multline}
  {\bf Level \ 1: \ \ } 
  \min_{{w_d},b,e_k} \mathcal{J}_\eta (w_d,e) =
  \frac{1}{2} \sum_{d=1}^D \frac{{w_d}^T w_d}{\eta_d} + \frac{1}{2}\sum_{k=1}^N e_k^2 \\
  \text{ \ s.t. \ }  \sum_{d=1}^D {w_d}^T \varphi_d(x^d_k) + b + e_k = y_k, \text{ \ \ } k=1,\dots,N.
  \label{eq.fusionc.cost}
\end{multline}
Note that the regularization vector appears here similar as in the 
Tikhonov regularization scheme \cite{tikhonov77} where each component 
is regularized individually. The Lagrangian of the constrained optimization problem 
with multipliers $\alpha^\eta \in\mathbb{R}^N$ becomes 
\begin{multline}
  \mathcal{L}_{\eta}(w_d,b,e_k;\alpha_k) 
  = \frac{1}{2} \sum_{d=1}^D \frac{{w_d}^T w_d}{\eta_d}  + \frac{1}{2} \sum_{k=1}^N e_k^2 \\
    - \sum_{k=1}^N \alpha^\eta_k ( \sum_{d=1}^D {w_d}^T \varphi_d(x_k^d) + b + e_k - y_k).
  \label{eq.fusionc.lag}
\end{multline}
By taking the conditions for optimality 
$\partial \mathcal{L}_\eta/\partial \alpha_k = 0$,  
$\partial \mathcal{L}_\eta/\partial b  = 0$,  
$\partial \mathcal{L}_\eta/\partial e_k = 0$ and
$\partial \mathcal{L}_\eta/\partial w_d = 0 $,
one gets the following conditions for optimality
\begin{equation}
  \left\{\begin{array}{rlrl}
      y_k &= \sum_{d=1}^D {w_d}^T \varphi_d(x_k^d) + b + e_k, & k=1,\dots,N & (a) \\
      e_k &= \alpha^\eta_k,                               & k=1,\dots,N & (b) \\
      w_d &= \eta_d\sum_{k=1}^N \alpha^\eta_k \varphi_d(x_k^d),        & d=1,\dots,D   & (c) \\
      0   &= \sum_{k=1}^N \alpha^\eta_k. && (d)
    \end{array}\right. 
  \label{eq.fusionc.cond}
\end{equation}
The dual problem is summarized in matrix notation 
by application of the kernel trick,  
\begin{equation}
  \left[
    \begin{tabular}{c|c}
      $0$ & $1_{N}^{T}$ \\ \hline
      $1_{N}$ & $\Omega^\eta +I_{N}$\\ 
    \end{tabular}
  \right]
  \left[ 
    \begin{tabular}{c}
      $b$ \\ \hline
      $\alpha^\eta$ \\ 
    \end{tabular}
  \right]
  = 
  \left[ 
    \begin{tabular}{c}
      $0$ \\ \hline
      $Y$ \\ 
    \end{tabular}
  \right],
  \label{eq.fusionc.train}
\end{equation}
where $\Omega^\eta \in \mathbb{R}^{N \times N}$ with 
$\Omega^\eta = \sum_{d=1}^D \eta_d\Omega^d$ and 
$\Omega^d_{kl} = K^d(x_k^d,x_l^d)$.
A new point $x_\ast \in \mathbb{R}^D$ can be evaluated as 
\begin{equation}
  \hat{y}_\ast = \hat{f}(x_\ast; \hat\alpha^\eta,\hat{b}) 
  = \sum_{k=1}^N \hat\alpha^\eta_k \sum_{d=1}^D \eta_d K^d(x_k^d,x_\ast^d) + \hat{b},
  \label{eq.fusionc.eval}
\end{equation}
where $\hat\alpha$ and $\hat{b}$ are the solution to (\ref{eq.fusionc.train}).
Simulating a training datapoint $x_k$ for all $k=1,\dots,N$ by the $d$-th individual component
\begin{equation}
  \hat{y}^{\eta,d}_k = \hat{f}^d(x_k^d; \hat\alpha^\eta) = \eta_d \sum_{l=1}^N \hat\alpha^\eta_l K^d(x_k^d,x_l^d), 
  \label{eq.fusionc.evald}
\end{equation}
which can be summarized in a vector $\hat{Y}^{\eta,d} = (\hat{y}^d_1,\dots,\hat{y}^d_N) \in \mathbb{R}^N$.
As in the previous section, the validation performance is used for tuning the regularization parameters
\begin{equation}
  {\bf\small Level \ 2: \ \ }
  \hat\eta  = \argmin_\eta \sum_{j=1}^n \left( f(x_j ;\hat\alpha^\eta,\hat{b}) - y_j\right)^2
  \mbox{ \ \ with \ \ } (\hat\alpha^\eta,\hat{b}) = \argmin_{\alpha^\eta,b} \mathcal{J}_\eta,
 \label{eq.fusionc.tuning}
\end{equation}
or using the conditions for optimality (\ref{eq.fusionc.train}) and eliminating $w$ and $e$ 
\begin{equation}
  {\bf\small Fusion: \ \ }
  (\hat\eta,\hat\alpha^\eta,\hat{b}) = \argmin_{\eta,\alpha^\eta,b} \sum_{j=1}^n \left( f(x_j ;\alpha^\eta,b) - y_j\right)^2
  \mbox{ \ \ s.t. \ \ (\ref{eq.fusionc.train}) holds},\hspace{+10mm}
  \label{eq.fusionc.fusion}
\end{equation}
which is a non-convex constrained optimization problem.

Embedding this problem in the additive regularization framework will lead us to a 
more suitable representation allowing for the use of dedicated algorithms. 
By relating the conditions (\ref{eq.areg.train}) to (\ref{eq.fusionc.train}), 
one can view the latter within the additive regularization framework by imposing extra 
constraints on $c$. The bias term $b$ is omitted from the remainder of this subsection for notational convenience.
The first two constraints reflect training conditions for both schemes. As the solutions 
$\alpha^\eta$ and $\alpha$ do not have the same meaning 
(at least for model evaluation purposes, see (\ref{eq.clssvmr.eval}) and (\ref{eq.fusionc.eval})), 
the appropriate $c$ is determined here by enforcing the same estimation on the training data. 
In summary:
\begin{equation}
  \renewcommand{\arraystretch}{1.5}
  \renewcommand{\arraycolsep}{8pt}
  \left\{\begin{array}{l}
      \left(\Omega + I_N\right)\alpha + c = Y\\
      \left(\left(\sum_{d=1}^D \eta_d\Omega^d\right) + I_N\right)\alpha^\eta  = Y\\
      \left(\sum_{d=1}^D \eta_d\Omega^d \right) \alpha^\eta = \Omega \alpha
  \end{array}\right.
  \Rightarrow
  \left\{\begin{array}{l}
      \left(\Omega + I_N\right)\alpha + c = Y\\
      \Omega \alpha = \eta^T \otimes I_N 
        \left[\begin{tabular}{c} $\Omega^1$\\$\hdots$\\$\Omega^D$ \\ \end{tabular}\right] 
        (\alpha + c),
  \end{array}\right.
  \label{eq.fusionc.c}
\end{equation}
where the second set of equations is obtained by eliminating $\alpha^\eta$.
The last equation of the righthand side represents the set of constraints of the values $c$ 
for all possible values of $\eta$.  The product $\otimes$ denotes 
$\eta^T \otimes I_N = [\eta_1 I_N, \dots, \eta_D I_N] \in\mathbb{R}^{N \times ND}$.
As for the Tikhonov case, it is readily seen that the solution space of $c$ 
with respect to $\eta$ is non-convex, however, the constraint on $c$ 
is recognized as a bilinear form. The fusion problem (\ref{eq.fusionc.fusion}) can be written as 
\begin{equation}
  {\bf\small Fusion: \ \ }
  (\hat\eta,\hat\alpha,\hat{c}) = \argmin_{\eta,\alpha,c} 
  \left\| \Omega^{(v)}\alpha - Y^{(v)}\right\|^2_2 
  \mbox{ \ \ s.t. \ \ (\ref{eq.fusionc.c}) holds},\hspace{+10mm}
  \label{eq.fusionc.fusion2}
\end{equation}
where algorithms as alternating least squares can be used.

%%%%%%%%%%%%%%%%%%%%%
\section{Applications}
%%%%%%%%%%%%%%%%%%%%%
\label{sect.ex}

For practical applications, the following iterative approach is used for solving non-convex cost-functions 
as (\ref{eq.aregam.cost2}). It can also be used for the efficient solution of convex
optimization problems which become computational heavy in the case of a large number of 
datapoints as e.g. (\ref{eq.aregam.costb}). A number of classification as well as 
regression problems are employed to illustrate the capabilities of the described 
approach. In the experiments, hyper-parameters as the kernel parameter (taken to be 
constants over the components) and the regularization trade-off parameter 
$\gamma$ or $\xi$ were tuned using $10$-fold cross-validation.

\subsection{Weighted Graduated Non-Convexity Algorithm}
%%%%%%%%%%%%%%%%%%%%%%%%%%%%%%%%%%%%%%%%%%%%%%%%%%%%%%
\label{subs.wgna}

An iterative scheme was developed based on the graduated non-convexity algorithm 
as proposed in \cite{blake89,nikolova99,antoniadis01} for the optimization of 
non-convex cost functions. 
%This kind of iterative approaches is similar to interior 
%point algorithms, see e.g. \cite{boyd04}. 
Instead of using a local gradient (or Newton) step which can be quite involved, %and assumes continuity 
an adaptive weighting scheme is proposed: in every step, the relaxed cost function is optimized by using 
a weighted 2-norm where the weighting terms are chosen based on an initial guess for the global solution. 
For every symmetric loss function $\ell(|e|):\mathbb{R}^+ \rightarrow \mathbb{R}^+$ 
which is monotonically increasing, there exists a bijective transformation $t:\mathbb{R} \rightarrow \mathbb{R}$ 
such that for every $e = y - f(x;\theta) \in \mathbb{R}$
\begin{equation}
  \ell(e) = \left( t(e) \right)^2.
  \label{eq.oracle.sim}
\end{equation}
The proposed algorithm for computing the solution for semi-norms
employs iteratively convex relaxations of the prescribed non-convex norm.
It is somewhat inspired by the simulated annealing optimization 
technique for optimizing global optimization problems.
The weighted version is based on the following derivation 
\begin{equation}
  \ell(e_k) = (\nu_k e_k)^2 \Leftrightarrow \nu_k = \sqrt{\frac{\ell(e_k)}{e^2_k}},
  \label{eq.norms}
\end{equation}
where the $e_k$ for all $k=1,\dots,N$ are the residuals corresponding with the 
solutions to $\theta = \argmin_\theta \ell\left(y_k - f(x_k;\theta)\right)$ 
\begin{figure}
  \centering
  \begin{tabular}{cc}\\
    \subfigure[]{\scalebox{.3}{\epsfig{figure=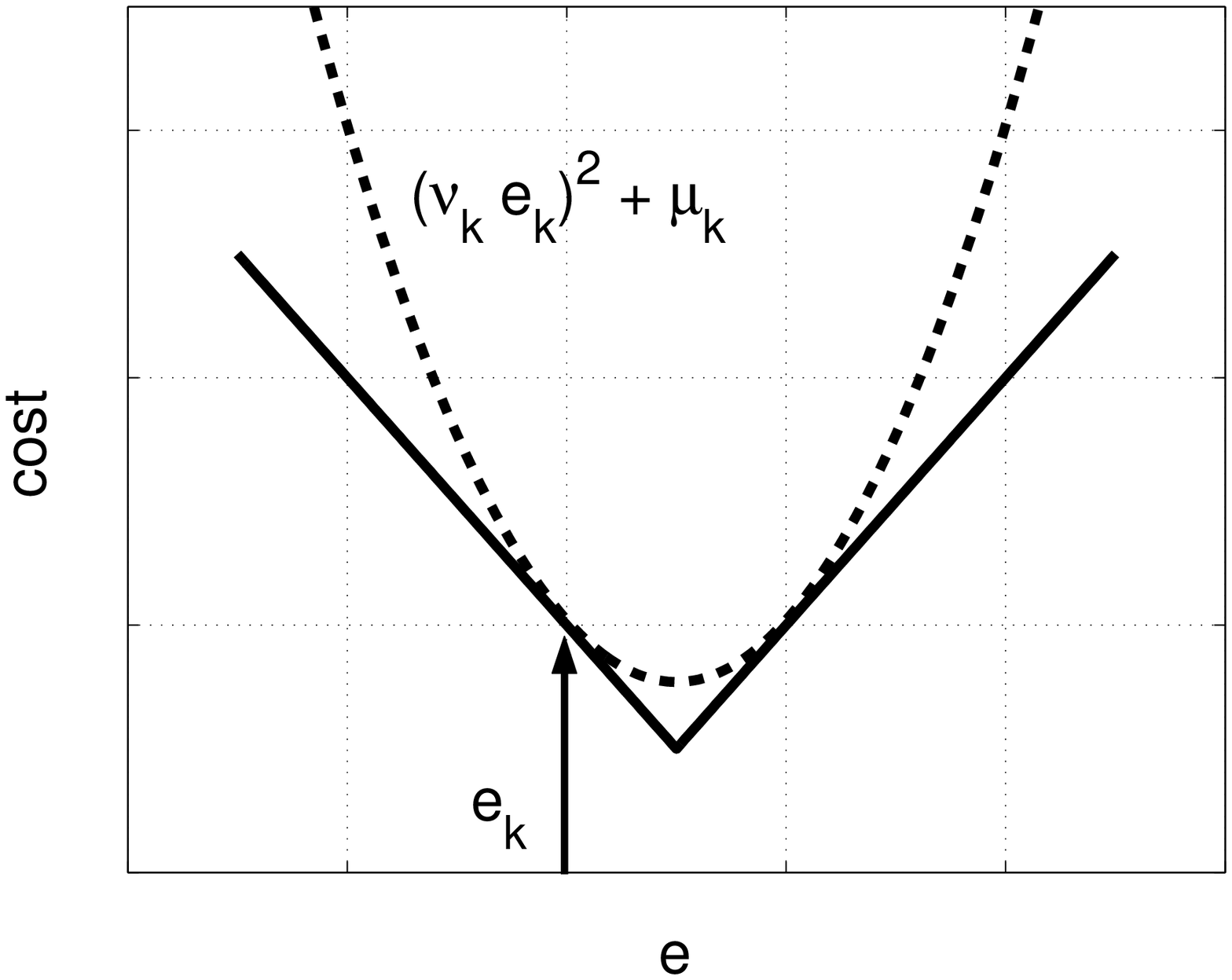}}} &
    \subfigure[]{\scalebox{.3}{\epsfig{figure=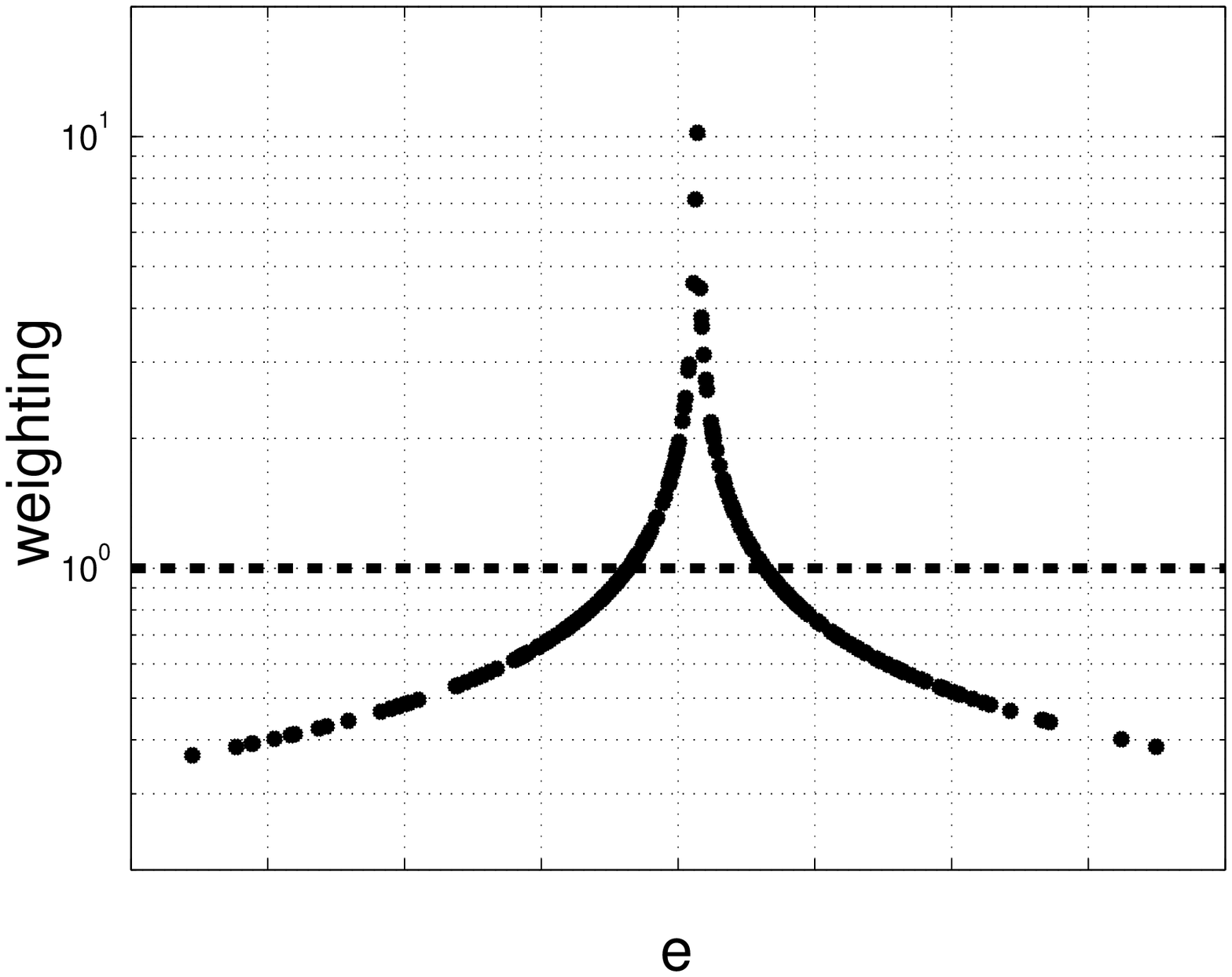}}}
  \end{tabular}
  \caption{\sl 
    {\bf (a)} Weighted $L_2$-norm (dashed) approximation $(\nu_k e_k)^2 + \mu_k$ 
    of the $L_1$-norm (solid) $\ell(e) = |e|_1$ which follows from the linear set of equations 
    (\ref{eq.norms2}) once the optimal $e_k$ are known; 
    {\bf (b)} the weighting terms $\nu_k$ for a sequence of $e_k$ and $k=1,\dots,N$
    such that $(\nu_k e_k)^2 + \mu_k = |e_k|_1$ and $2\nu_k^2 e_k = l'(e_k) = {\rm sign}(e_k)$
    for an appropriate $\mu_k$.}
  \label{fig.it}
\end{figure}
This is equal to the solution of the convex optimization problem 
$e_k = \argmin_\theta \left(\nu_k (y_k - f(x_k;\theta))\right)^2$ 
for a set of $\nu_k$ satisfying (\ref{eq.norms}).
For more stable results, the gradient of the penalty function 
$\ell$ and the quadratic approximation can be takne equal as follows
by using an intercept parameter $\mu_k \in\mathbb{R}$ for all $k=1,\dots,N$:
\begin{equation}
  \left\{\begin{array}{rl}
      \ell(e_k) &= (\nu_k e_k)^2 + \mu_k \\
      \ell'(e_k)&=  2 \nu^2_k e_k 
  \end{array}\right.
\Leftrightarrow 
  \left[\begin{tabular}{c|c}$e^2_k$&$1$\\\hline $2e_k$&$0$\\\end{tabular}\right]
  \left[\begin{tabular}{c}$\nu_k^2$\\\hline$\mu_k$\\\end{tabular}\right]
  =
  \left[\begin{tabular}{c}$\ell(e_k)$\\\hline$\ell'(e_k)$\\\end{tabular}\right],
  \label{eq.norms2}
\end{equation}
where $\ell'(e_k)$ denotes the derivative of $\ell$ evaluated in $e_k$ such that 
a minimum of $\mathcal{J}_\ell$ also minimizes the weighted equivalent (the derivatives are equal). 
Note that the constant intercepts $\mu_k$ are not relevant in the weighted optimization problem. 
Under the assumption that the two consecutive relaxations $\ell^{(t)}$ and $\ell^{(t+1)}$
do not have too different global solutions, the following algorithm is a plausible practical tool:
\begin{Algorithm}[Weighted Graduated Non-Convexity Algorithm]
  For the optimization of semi-norms ($\ell(\cdot)$), a practical approach is based on 
  deforming gradually a 2-norm into the specific loss function of interest.
  Let $\zeta$ be a strictly decreasing series $1,\zeta^{(1)},\zeta^{(2)},\dots,0$.
  A plausible choice for the initial convex cost function is the least squares 
  cost function $\mathcal{J}_{\rm LS}(e) = \|e\|_2^2$. 
  \begin{enumerate}
  \item Compute the solution $\theta^{(0)}$ for $L_2$ norm $J_{\rm LS}(e) = \|e\|_2^2$ 
    with residuals $e_k^{(0)}$;
  \item $t = 0$ and $\nu^{(0)} = 1_N$;
  \item Consider the following relaxed cost function 
    $\mathcal{J}^{(t)}(e) = (1-\zeta_t) \ell(e) + \zeta_t \mathcal{J}_{\rm LS}(e)$;
  \item Estimate the solution $\theta^{(t+1)}$ and corresponding residuals $e_k^{(t+1)}$ 
    of the cost function $\mathcal{J}^{(t)}$ using the weighted approximation 
    $\mathcal{J}_{\rm approx} = (\nu_k^{(t)} e_k)^2$ of $J^{(t)}(e_k)$
  \item Reweight the residuals using weighted approximative squares norms 
    as derived in (\ref{eq.norms2}):
  \item $t := t+1$ and iterate step (3,4,5,6) until convergence.
  \end{enumerate}
\end{Algorithm}
When iterating this scheme, most $\nu_k$ will be smaller than $1$ as the least squares cost function
penalizes higher residuals (typically outliers). However, a number of residuals will have increasing 
weight as the least squares loss function is much lower for small residuals.

\begin{figure}
   \begin{center}
     \scalebox{.6}{\epsfig{figure=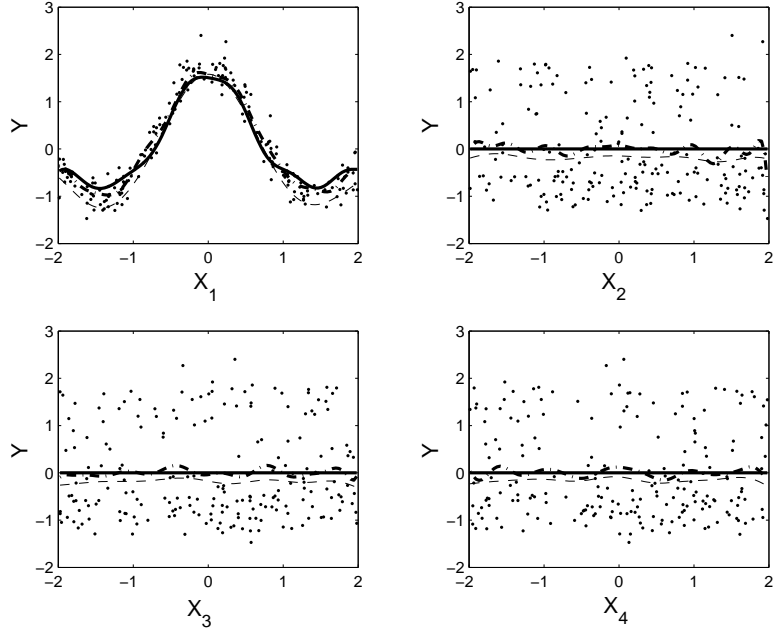}}
   \end{center}
   \caption{\sl Example of a toy dataset consisting of four input components $X^1,X^2,X^3$ and $X^4$ 
     where only the first one is relevant to predict the output $f(x) = {\rm sinc}(x^1)$. 
     A componentwise LS-SVM regressor (dashed line) has good prediction performance, while 
     the $L_1$ penalized cost function of Subsection (\ref{subs.1norm}) also recovers the 
     structure in the data as the estimated components correspnding with $X^2,X^3$ and $X^4$ are sparse. }
   \label{fig.sinc}
 \end{figure}

\subsection{Regression examples}
%%%%%%%%%%%%%%%%%%%%%%%%%%%%%%%

To illustrate the additive model estimation method, a classical example was constructed as in \cite{hastie90,vapnik98}.
The data were generated according to $y_k = 10\,{\rm sinc}(x^1_k) +  20\,(x^2_k-0.5)^2 + 10\, x^3_k + 5\, x^4_k + e_k$
were $e_k \sim \mathcal{N}(0,1)$, $N = 100$ and the input data $X$ are randomly chosen from the interval $[0,1]^{10}$.
Because of the Gaussian nature of the noise model, only results from least squares methods are reported.
The described techniques were applied on this training dataset and tested on an independent test set generated using the saem rules. 
Table \ref{table.ex} reports whether the algorithm recovered the structure in the data (if so, the measure is 100\%).
The experiment using the smoothly tresholding penalized (STP) cost function was designed as follows: for 
every 10 components, a version was provided for the algorithm for the use of a linear kernel and another for the 
use of a RBF kernel (resulting in 20 new components). The regularization scheme was able to select the components with the appropriate
kernel (a nonlinear RBF kernel for $X^1$ and $X^2$ and linear ones for $X^3$ and $X^4$), 
except for one spurious component (A RBF kernel was selected for the fifth component).

\begin{table}
  \renewcommand{\arraystretch}{1.5}
  \renewcommand{\arraycolsep}{15pt}
  \centering
  \begin{tabular}{|c||c|c|c|c|}\hline
    {\bf Method} & \multicolumn{3}{c|}{\bf Test set Performance} & {\bf Sparse components} \\ 
                 & $L_2$ & $L_1$ & $L_\infty$ & \% recovered \\ \hline\hline
    LS-SVMs      & 0.1110 & 0.2582 & 0.8743 & 0\% \\
    componentwise LS-SVMs (\ref{eq.clssvmr.train})
                 & 0.0603 & 0.1923 & 0.6249 & 0\% \\
    $L_1$ regularization (\ref{eq.aregam.costb})
                 & 0.0624 & 0.1987 & 0.6601 & 100\%\\ 
    STP with RBF (\ref{eq.aregam.cost2})        
                 & 0.0608 & 0.1966 & 0.6854 & 100\% \\
    STP with RBF and lin (\ref{eq.aregam.cost2})
                 & 0.0521 & 0.1817 & 0.5729 & 95\% \\
    Fusion with AReg (\ref{eq.fusion1.cost})     
                 & 0.0614 & 0.1994 & 0.6634 & 100\% \\
    Fusion with comp. reg. (\ref{eq.fusionc.fusion2}) 
                 & 0.0601 & 0.1953 & 0.6791 & 100\% \\ \hline
  \end{tabular}
  \caption{\sl Results on test data of numerical experiments on the Vapnik regression dataset.
    The sparseness is expressed in the rate of components which is selected only if the input is relevant
    (100\% means the original structure was perfectly recovered).}
  \label{table.ex}
\end{table}

\subsection{Classification example}
%%%%%%%%%%%%%%%%%%%%%%%%%%%%%%%%%%%

\begin{figure}
   \begin{center}
     \scalebox{.7}{\epsfig{figure=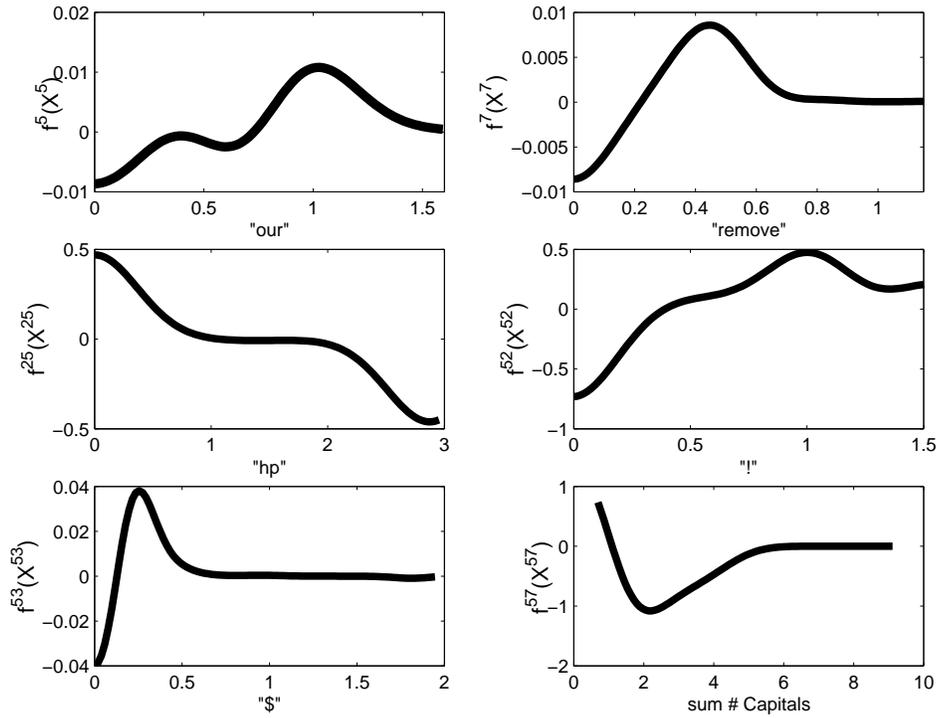}}
   \end{center}
   \caption{\sl Results of the spam dataset. The non-sparse components as found by application 
     of Subsection \ref{subs.oracle} are shown suggesting a number of usefull indicator variables 
     for classifing a mail message as spam or non-spam. The final classifier takes the form 
     $f(X) = f^5(X^5) + f^{7}(X^{7}) + f^{25}(X^{25})+ f^{52}(X^{52})+ f^{53}(X^{53})+ f^{56}(X^{56})$
     where 6 relevant components were selected out of the 56 provided indicators.}
   \label{fig.spam}
 \end{figure}

An additive model was estimated by an LS-SVM classifier based on the \verb|spam| data
as provided on the UCI benchmark repository, see e.g. \cite{hastie01}. The data consists of 
word frequencies from 4601 email messages, in a study to screen email for spam.
A test set of size 1536 was drawn randomly from the data leaving 3065 to training purposes.
The inputs were preprocessed using following transformation $p(x) = {\rm log}(1+x)$ and 
standardized to unit variance.
Figure \ref{fig.spam} gives the indicator functions as found using a regularization based 
technique to detect structure as described in Subsection \ref{subs.oracle}.
The structure detection algorithm selected only 6 out of the 56 provided 
indicators. Moreover, the componentwise approach describes the form of the contribution
of each indicator, resulting in an highly interpretable model.

%%%%%%%%%%%%%%%%%%%%%
\section{Conclusions}
%%%%%%%%%%%%%%%%%%%%%
\label{sect.concl}

This chapter describes nonlinear additive models based on LS-SVMs which are 
capable of handling higher dimensional data for regression as well as 
classification tasks. The estimation stage results from solving a set 
of linear equations with a size approximatively equal to the number of training datapoints. 
Furthermore, the additive regularization framework is employed for 
formulating dedicated regularization schemes leading to structure detection.
Finally, a fusion argument for component selection and structure detection
based on training componentwise LS-SVMs and validation performance is introduced 
to improve the generalization abilities of the method.
Advantages of using componentwise LS-SVMs include the 
efficient estimation of additive models with respect to classical 
practice, interpretability of the estimated model, opportunities  
towards structure detection and the connection with existing statistical techniques.

\hspace{+5mm}

{\footnotesize {\bf Acknowledgments}.
This research work was carried out at the ESAT laboratory of the Katholieke Universiteit Leuven.
Research Council KUL: GOA-Mefisto 666, GOA AMBioRICS, several PhD/postdoc \& fellow grants;
Flemish Government:
FWO: PhD/postdoc grants, projects, G.0240.99 (multilinear algebra), G.0407.02 (support vector machines), G.0197.02 (power islands), G.0141.03 (Identification and cryptography), G.0491.03 (control for intensive care glycemia), G.0120.03 (QIT), G.0452.04 (new quantum algorithms), G.0499.04 (Robust SVM), G.0499.04 (Statistics) research communities (ICCoS, ANMMM, MLDM);
AWI: Bil. Int. Collaboration Hungary/ Poland;
IWT: PhD Grants,GBOU (McKnow)
Belgian Federal Science Policy Office: IUAP P5/22 (`Dynamical Systems and Control: Computation, Identification and Modelling', 2002-2006) ; PODO-II (CP/40: TMS and Sustainability);
EU: FP5-Quprodis; ERNSI; Eureka 2063-IMPACT; Eureka 2419-FliTE;
Contract Research/agreements: ISMC/IPCOS, Data4s, TML, Elia, LMS, Mastercard
is supported by grants from several funding agencies and sources.
GOA-Ambiorics, IUAP V, FWO project  G.0407.02 (support vector machines)
FWO project  G.0499.04 (robust statistics)
FWO project  G.0211.05 (nonlinear identification)
FWO project  G.0080.01 (collective behaviour)
JS is an associate professor and BDM is a full professor at K.U.Leuven Belgium, respectively.}

\bibliographystyle{ifac}
\bibliography{refs}

\printindex
\end{document}